\documentclass{article}
\usepackage{spconf,amsmath,graphicx}
\usepackage{algorithmic}
\usepackage{algorithm}
\usepackage{hyperref}
\usepackage{caption}
\usepackage{subcaption}

\title{Neural Topic Modeling of Psychotherapy Sessions}
\name{Baihan Lin$^1$, Djallel Bouneffouf$^2$, Guillermo Cecchi$^2$,  Ravi Tejwani$^3$}
\address{
  $^1$Columbia University\\
  $^2$IBM Thomas J. Watson Research Center \\
  $^3$Massachusetts Institute of Technology \\
{baihan.lin@columbia.edu, gcecchi@us.ibm.com, djallel.bouneffouf@ibm.com, tejwanir@mit.edu}}

\begin{document}
%
\maketitle
\begin{abstract}

In this work, we compare different neural topic modeling methods in learning the topical propensities of different psychiatric conditions from the psychotherapy session transcripts parsed from speech recordings. We also incorporate temporal modeling to put this additional interpretability to action by parsing out topic similarities as a time series in a turn-level resolution. We believe this topic modeling framework can offer interpretable insights for the therapist to optimally decide his or her strategy and improve psychotherapy effectiveness.

\end{abstract}

\section{Introduction}
Mental health remains an issue in all countries and cultures across the globe.  According to the National Institute of Mental Health (NIMH), nearly one in five U.S. adults live with a mental illness (52.9 million in 2020).
One of the major causes of the mental illness is depression \cite{li2022changes}, followed by suicide which is the second cause of death among young people \cite{moe2022risk}.  It is clear that there is a need for new innovative solutions in this domain.
Psychotherapy is a term given for treating mental health problems by talking with a mental health provider such as a psychiatrist or psychologist\cite{ahmad2022designing}.
To reduce the workload on mental health provider, natural language processing (NLP) is more and more adopted \cite{rezaii2022natural}. Noting that psychotherapy has been the first discipline using NLP. It started with a chat bot ELIZA\cite{shum2018eliza} capable of mimicking a psychotherapist. Another chatbot, Parry \cite{zemvcik2019brief}, was able of simulating an individual with Schizophrenia. 
Natural language processing including topic modeling has shown interesting results on mental illness detection. In \cite{resnik2015beyond} the authors demonstrate that Latent Dirichlet Allocation (LDA) can uncover latent structure within depression-related language collected from Twitter. Authors \cite{zeng2012synonym} shows the add-value of using social media content to detect Post-Traumatic Stress Disorder. 

Although previous works demonstrate the effectiveness of classical topic modeling, they are no longer the state-of-the-art.
In recent years, deep learning progresses the fields and the Neural Topic Modeling shows up as the consistent better solution compared to the classical Topic modeling \cite{miao2016neural}. In this context, we propose in this work to use Neural Topic Modeling to learn the topical propensities of different psychiatric conditions from the psychotherapy session transcripts. We benchmark our findings on the Alex Street Counseling and Psychotherapy Transcripts dataset\footnote{https://alexanderstreet.com/products/counseling-and-psychotherapy-transcripts-series}, which consists of the transcribed recordings of over 950 therapy sessions between multiple anonymized therapists and patients with anxiety, depression, schizophrenia or suicidal intents. This multi-part collection includes speech-translated transcripts of the recordings from real therapy sessions, 40,000 pages of client narratives, and 25,000 pages of reference works.
In total, these materials include over 200,000 turns together for the patient and therapist and provide access to the broadest range of clients for our linguistic analysis of the therapeutic process of psychotherapy.

The goal here is to evaluate the existing techniques on neural topic modeling and find the most adapted one to this domain. Second, we incorporate temporal modeling to put additional interpretability, where the goal of this framework is to offer interpretable insights for the therapist to optimally decide on psychotherapy strategy.

\begin{figure}[tb]
\centering
    \includegraphics[width=\linewidth]{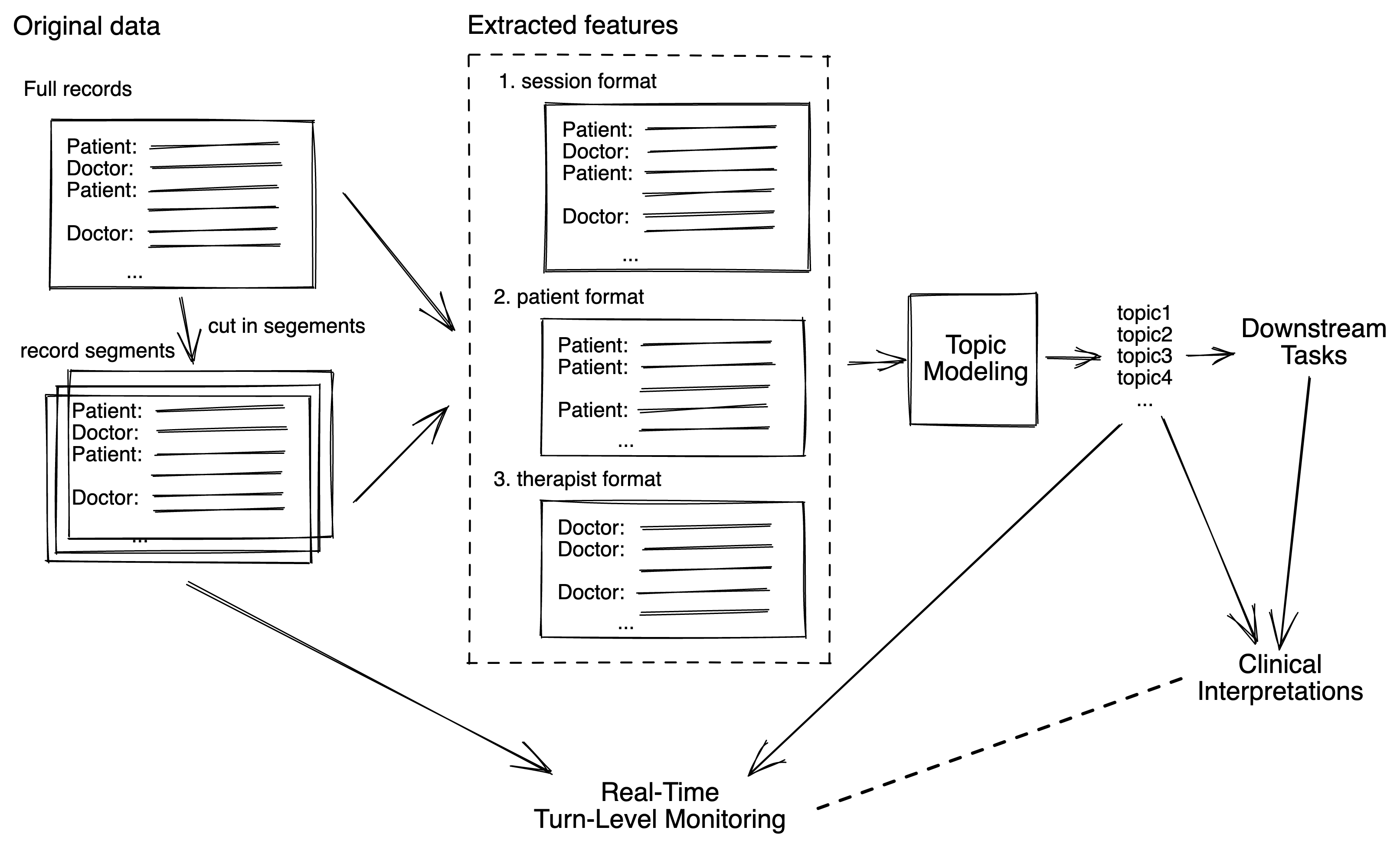}
\caption{Psychotherapy topic modeling framework
}\label{fig:pipeline}
\end{figure}

\section{Related Work on Topic Modeling}

In natural language processing and machine learning, a topic model is a type of statistical graphical model that help uncover the abstract ``topics'' that appear in a collection of documents. The topic modeling technique is frequently used in text-mining pipeline to unravel the hidden semantic structures of a text body. 
There are quite a few neural topic models evaluated in this work.
The Neural Variational Document Model (NVDM) \cite{miao2016neural} is an unsupervised text modeling approach based on variational auto-encoder. \cite{miao2017discovering} further shows that among NVDM variants, the Gaussian softmax construction (GSM) achieves the lowest perplexity in most cases, and thus recommended. We denote it as NVDM-GSM. Unlike traditional variational autoencoder based methods, Wasserstein-based Topic Model (WTM) uses the Wasserstein autoencoders (WAE) to directly enforce Dirichlet prior on the latent document-topic vectors \cite{nan2019topic}. Traditionally, it applies a suitable kernel in minimizing the Maximum Mean Discrepancy (MMD) to perform distribution matching. We can this variant WTM-MMD. Similarly, we can replace the MMD priors with a Gaussian Mixture prior and apply Gaussian Softmax on top of it. We denote this method, WTM-GMM. In order to tackle the issue with large and heavy-tailed vocabularies, the Embedded Topic Model (ETM) \cite{dieng2020topic} models each word with a matched categorical probablity distribution given the inner product between a word embedding and a vector embedding  of its assigned topic. To avoid imposing impropoer priors, Bidirectional Adversarial Training Model (BATM) applies the bidirectional adversarial training into neural topic modeling by constructing a two-way projection between the document-word distribution and the document-topic distribution \cite{wang2020neural}.


\section{Therapy topic modeling framework}


Figure \ref{fig:pipeline} is an outline of the analytic framework. During the session, the dialogue between the patient and therapist are transcribed into pairs of turns. 
We take the full records of a patient, or a cohort of patients belonging to the same condition. We either use it as is before the feature extraction, or we truncate them into segments based on timestamps or topic turns. 
When we have the features, we fit them into the topic models. The end results of the topic modeling would be a list of weighted topic words, that tells us what the text block is concerned with. These knowledges are usually very informative and interpretable, thus important in psychotherapy.

Here are a few downstream tasks and user scenarios that can plugged to our analytical frameworks. We can either use these extracted weighted topics to inform whether the therapy is going the right direction, whether the patient is going into certain bad mental state, or whether the therapist should adjust his or her treatment strategies. This can be built as an intelligent AI assistant to remind the therapist of such things. Some topics can also be off-limit taboos, such as those in suicidal conversations, so if such terms arises from the topic modeling (say, a dynamic topic modeling), it can be flagged for the doctor to notice.


\begin{algorithm}[H]
 \caption{Temporal Topic Modeling (TTM)}
 \label{alg:ttm}
 \begin{algorithmic}[1]
 \STATE {\bfseries } Learned topics $T$ as references
 \STATE {\bfseries }\textbf{for} i = 1,2,$\cdots$, N \textbf{do}
\STATE {\bfseries } \quad Automatically transcribe dialogue turn pairs  $(S^p_i,S^t_i)$
\STATE {\bfseries }\quad \textbf{for} $T_j \in$ topics $T$ \textbf{do} \\
\STATE {\bfseries }\quad \quad Topic score $W^{p_i}_{j}$ = similarity($Emb({T_j}), Emb(S^p_i)$) \\
\STATE {\bfseries }\quad \quad Topic score $W^{t_i}_{j}$ = similarity($Emb({T_j}), Emb(S^t_i)$) \\
\STATE {\bfseries } \quad \textbf{end for}
\STATE {\bfseries } \textbf{end for}
 \end{algorithmic}
\end{algorithm}

Given the learned topics, we can backtrack the transcript to get a turn-resolution topic scores. Algorithm \ref{alg:ttm} outlines the pipeline of our temporal topic modeling analysis (TMM). Say, if we have learned 10 topics, the topic score will be a vector of 10 dimensions, with each dimension corresponding to some notion of likelihood of this turn being in this topic. Because we want to characterize the directional property of each turn with a certain topic, we compute the cosine similarity of the embedded topic vector and the embedded turn vector, instead of directly inferring the probability as traditional topic assignment problem (which would be more suitable if we merely want to find the assignment of the most likely topic). In the result section, we will present the temporal modeling of the Embedded Topic Model (ETM), but this analytic pipeline can in principle be applied to any learned topic models. This Embedded Topic Model is special because, like our approach here, it also models each word with a categorical distribution whose natural parameter is the inner product between a word embedding and an embedding of its assigned topic. We use the same word embedding here (Word2Vec \cite{mikolov2013distributed}) to embed our topic and turns.

 

\section{Results}

\begin{table*}[t]
      \caption{Coherence embedding evaluations of the neural topic models (following \cite{roder2015exploring})
      }
      \label{tab:emb_eval} 
      \centering
      \resizebox{0.95\linewidth}{!}{
 \begin{tabular}{ l | c | c | c | c | c | c | c | c | c | c | c | c | }
 &\multicolumn{4}{c}{Anxiety} \vline &\multicolumn{4}{c}{Depression} \vline &\multicolumn{4}{c}{Schizophrenia} \vline \\
  & $c_v$ & $c_{w2v}$ & $c_{uci}$ & $c_{npmi}$   & $c_v$ & $c_{w2v}$ & $c_{uci}$ & $c_{npmi}$   & $c_v$ & $c_{w2v}$ & $c_{uci}$ & $c_{npmi}$  \\ \hline
NVDM-GSM & 0.410 & 0.484 & -0.844 & -0.019 & 0.495 & 0.531 & -3.522 & -0.109 & 0.642 & - & -1.954 & -0.065 \\
WTM-MMD & 0.340 & 0.428 & -2.827 & -0.099 & 0.290 & 0.462 & -3.797 & -0.124 & 0.576 & 0.751 & -0.997 & -0.036 \\
WTM-GMM & 0.353 & 0.413 & -3.259 & -0.116 & 0.678 & 0.535 & -0.126 & -0.006  & 0.572 & 0.774 & -1.587 & -0.050 \\
ETM & 0.413 & - & -2.903 & -0.093 & 0.403 & - & -2.399 & -0.05 & 0.379 & 0.864 & -7.232 & -0.199 \\
BATM & 0.352 & 0.387 & -5.056 & -0.190 & 0.404 & 0.423 & -4.238 & -0.160 & 0.507 & 0.816 & -9.655 & -0.343 \\
\end{tabular}
 }
\end{table*}

\begin{table*}[t]
      \caption{Topic evaluations of the neural topic models (following \cite{mimno2011optimizing})
      }
      \label{tab:eval} 
      \centering
      \resizebox{0.95\linewidth}{!}{
 \begin{tabular}{ l | c | c | c | c | c | c |  }
 &\multicolumn{2}{c}{Anxiety} \vline &\multicolumn{2}{c}{Depression} \vline &\multicolumn{2}{c}{Schizophrenia} \vline \\
  & Topic coherence & Topic diversity  & Topic coherence & Topic diversity   & Topic coherence & Topic diversity   \\ \hline
NVDM-GSM & 0.653 & \textbf{-380.933} & 0.487 & -316.439 & 0.527 & -431.393 \\
WTM-MMD & \textbf{0.927} & -453.929 & 0.907 & -359.964 & 0.447 & -403.694 \\
WTM-GMM & 0.907 & -425.515 & 0.340 & \textbf{-236.815} & 0.467 & \textbf{-204.930} \\
ETM & 0.893 & -449.000 & \textbf{0.933} & -367.069 &  \textbf{0.973} & -310.211 \\
BATM & 0.720 & -441.049 & 0.773 & -443.394 & 0.500 & -337.825 \\
\end{tabular}
 }
\end{table*}

In this section, we compare five state-of-the-art neural topic modeling approaches introduced above, and analyze their learned topics. We separate transcript sessions into three categories based on the psychiatric conditions of the patients (anxiety, depression and schizophrenia), and train the topic models over each of them for over 100 epochs at a batch size of 16. As in the standard preprocessing of topic modeling training, we set the lower bound of count for words to keep in topic training to be 3, and the ratio of upper bound of count for words to keep in topic training to be 0.3. The evaluation procedure follows the same implementation for \cite{wang2020neural}\footnote{https://github.com/zll17/Neural\_Topic\_Models}.  

\subsection{Evaluation metrics}

Topic models are usually evaluated with the likelihood of held-out documents and topic coherence. However, it was shown that a higher likelihood of held-out documents does not necessarily correlate to the human judgment of topic coherence \cite{chang2009reading}. Therefore, we adopt a series of more validated measurements of topic coherence and diversity by following \cite{roder2015exploring}. In the first evaluation, we compute four topic embedding coherence metrics ($c_{v}$, $c_{w2v}$, $c_{uci}$, $c_{npmi}$) to evaluate the topics generated by various models (as outlined in \cite{roder2015exploring}). The higher these measurements, the better. In all experiments, each topic is represented by the top 10 words according to the topic-word probabilities, and the four metrics are calculated using Gensim library \cite{vrehuuvrek2011gensim}\footnote{https://github.com/RaRe-Technologies/gensim}.
Other than these four topic embedding coherence evaluation provided by Gensim, we also included two other useful metrics. \cite{mimno2011optimizing} proposed a robust and automated coherencce evaluation metric for identifying such topics that does not rely on either additional human annotations or reference collections outside the training set. This method computes an asymmetrical confirmation measure between top word pairs (smoothed conditional probability). In addition, we compute the topic diversity by taking the ratio between the size of vocabulary in the topic words and the total number of words in the topics. Similarly, the higher these two measures are, the better the topic models.

\subsection{Quantitative evaluations of the topic models}

Tables \ref{tab:emb_eval} and \ref{tab:eval} summarize quantitative evaluations. We first observe that the different measures of the coherence gives different rankings of the topic models, but there are a few models that perform relatively well across the metrics.  WTM and ETM both yield relatively high topic coherence and diversity. 




 \begin{figure}[tb]
\centering
\includegraphics[width=.47\linewidth]{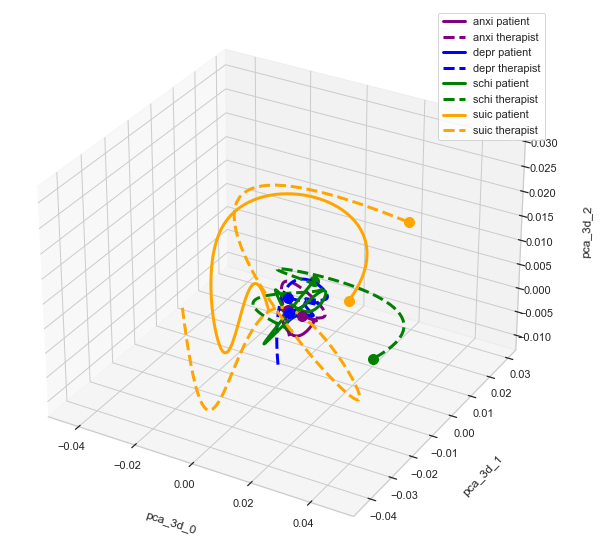}\hfill
    \includegraphics[width=.47\linewidth]{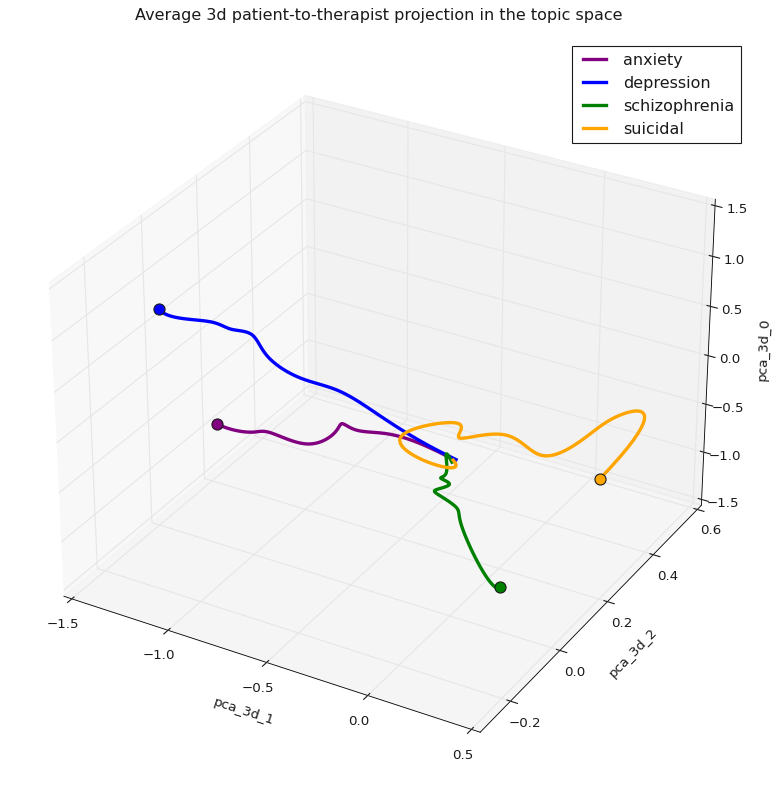}
\caption{The average 3d trajectories and the patient-to-therapist projections of different classes of psychiatric conditions in the principal topic space (dots are the trajectory end points).
}\label{fig:trajs_pca}
\end{figure}

\subsection{Temporal dynamics of the topic models}

To ensure that the topics can be mapped from one clinical condition to another condition, we compute a universal topic model on the text corpus of the entire Alex Street psychotherapy database. 
And then, given the learned topics from this universal topic models, we can compute a 10-dimensional topic score for each turn corresponding to the 10 topics. The higher the score is, the more positively correlated this turn is with this topic. Given this time-series matrix, we can potentially probe the dynamics of the these dialogues within the topic space. 
We can also provide more distinctive features for downstream tasks by performing a principal component analysis on the topic space. Figure \ref{fig:trajs_pca} presents the average temporal trajectories of the patients and therapists, as well as the patient-to-therapist projection (i.e. the vector difference in the patient-therapist pair) in the principal topic spaces. We observe that the suicidal sessions cover a wider variety of topics (by having more spread-out trajectories), 
and have a more curved patient-therapist topic difference with multiple twists along the full session, while the other three clinical conditions have more consistent directions of such differences. This might suggest that a strategy by the therapist to divert from the sensitive topics. In schizophrenia sessions, the therapist appears to cover a bigger topical arc than the patient, suggesting a therapeutic strategy of visiting multiple topics to distract the patient from sensitive ones. The topical trajectories of the anxiety and depression sessions, comparatively, are more converged. This is a first step of identifying the prototypical therapeutic strategies in different psychiatric conditions and a potential turn-level resolution temporal analysis of topic modeling. With this approach one can go over the sessions (as a therapist) and analyze the dynamics afterwards.
 

\subsection{Interpretable insights from the learned topics}

To provide interpretable insights, it is important to parse out the concepts behind these learned topics. To better understand what these topics are, we parse out the highest scoring turns in the transcripts that correspond to each topics. 

First, we dive into the individual topic models trained on text corpus of each psychiatric condition separately. For instance, here are the interpretations from the top scoring turns in the anxiety sessions: topic 0 is chit-chat and interjections; topic 1 is low-energy exercises; topic 2 is fear; topic 3 is medication planning; topic 4 is the past, control and worry; topic 5 is other people and some objects; topic 6 is just well being; topic 7 is music, headache and emotion; topic 8 is stress; and topic 9 is fear and responsibilities.
For depression, topic 0 is time; topic 1 is husband and anger; topic 2 is time and distance; topic 3 is energy and stress levels; topic 4 is self-esteem; topic 5 is money and time; topic 6 is age and time; topic 7 is mood and time; topic 8 is people and objects; topic 9 is holidays and chit-chats.
For schizophrenia, topic 0 is family; topic 1 is extreme terms; 
topic 2 is energy level and positives; topic 3 is people and family; topic 4 is operational stuffs; topic 5 is calm things;
topic 6 and 9 are critical topics.


For the universal topic models, the results are much more coherent. 
For instance, topic 0 is about figuring out, self-discovery and reminiscence. Topic 1 is about play. Topic 2 is about anger, scare and sadness. Topic 3 is about counts. Topic 4 is about tiredness and decision. Topic 5 is about sickness, self injuries and coping mechanisms. Topic 6 is about explicit ways to deal with stress, such as keep busying and reaching out for help. Topic 7 is about numbers. Topic 8 is about continuation and keep doing. Topic 9 is mostly chit-chat, interjections and transcribed prosody.

We notice that among all the clinical conditions we compare, the learned topics obtain a relatively poor mapping in the dialogue of suicidal cases. This might be due to the small sample size available in suicidal sessions, or the frequent hand annotations of behaviors (e.g. ``patient crying for a few minutes'' or ``patient leaves the room'') with time stamps, which doesn't conform to the annotation style of other sessions.

\subsection{Ranked topics informed by working alliance}

 \begin{figure}[tb]
\centering
\vspace{-0.5em}
\includegraphics[width=.83\linewidth]{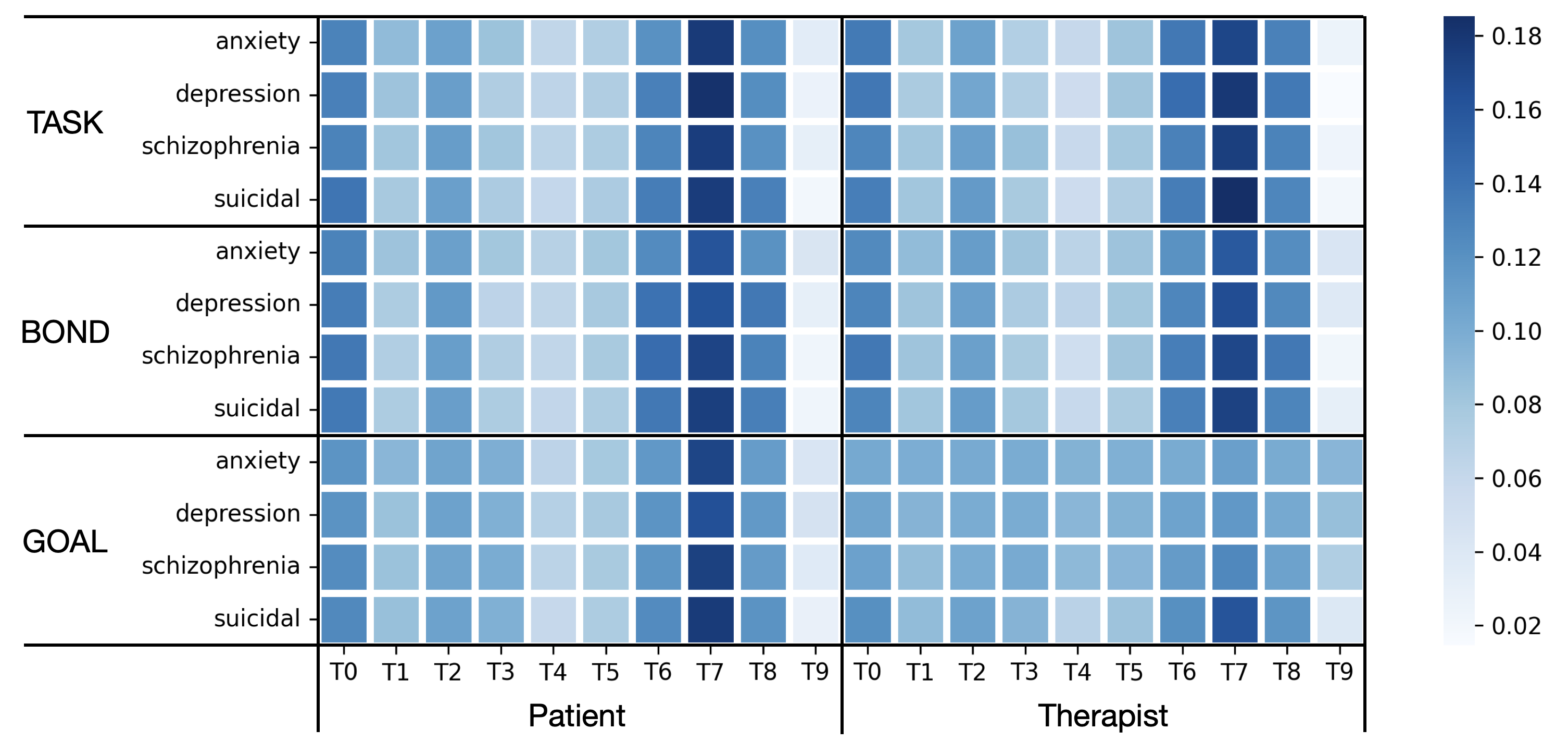}
\caption{Topic distributions of turns with top working alliance.
}\label{fig:wa_filter}
\vspace{-0.5em}
\end{figure}

Although our approach can annotate the topics in each dialogue turns of the psychotherapy sessions, we don't know how informative they might be from the therapeutic point of view. In \cite{lin2022deep}, we propose a computational technique to directly infer the therapeutic working alliance of a dialogue turn, which can be predictive of how effective the current therapy treatment is to the given patient at the given state. Combining this method with our topic modeling framework would enable us to highlight disorder-specific topics and dialogue segments that are potentially indicative of the therapeutic breakthroughs. For each disorder, we filter the turns to the top 100 working alliance scores, separately in three scales (task, bond and goal). Figure \ref{fig:wa_filter} is the heatmap of the averaged topic scores. We first observe no clear distinction among the working alliance scales, but notice a relatively uniform coverage of the topics when the patient and therapist are well aligned in the goal scale in all clinical conditions except for suicidal cases. Inspecting the top 10 turns with the highest topic scores, we notice that within the turns with high working alliance goal scale, suicidal patients tend to discuss sensitive terms like ``alive'', ``stop'' and ``sexual''.

\section{Conclusions}
In this work, our first goal is to compare different neural topic modeling methods in learning the topical propensities of different psychiatric conditions. 
We first observe that different measures of the coherence gives different rankings of the topic models, but there are a few topic models that perform relatively well across metrics. For instance, Wasserstein Topic Models and Embedded Topic Models both yield relatively high topic coherence and diversity. 
Our second goal is to parse topics in different segments of the session, which allows us to incorporate temporal modeling and add additional interpretability. For instance, these allows us to notice that the session trajectories of the patient and therapist are more separable from one another in anxiety and depression sessions, but more entangled in the schizophrenia sessions. This is the first step of a potential turn-level resolution temporal analysis of topic modeling. We believe this topic modeling framework can offer interpretable insights for the therapist to improve the psychotherapy effectiveness.

Next steps include predicting these topic scores as states (such as \cite{lin2020predicting,lin2022predicting}), training text or speech-based chatbots as reinforcement learning agents (as reviewed in \cite{lin2022reinforcement}) given these psychological or therapeutic states incorporating biological and cognitive priors (as in \cite{lin2020story,lin2020unified,lin2019split,lin2021models}) and studying its factorial relations with other inference anchors (e.g. working alliance and personality \cite{lin2022deep,lin2022deep2,lin2022unsupervised}). The end goal is to construct a complete AI knowledge management system of mental health utilizing different NLP annotations in real time, as outlined in this database-perspective review \cite{lin2022knowledge}.



\bibliographystyle{IEEEbib}
\bibliography{main}

\begin{thebibliography}{10}

\bibitem{li2022changes}
Joyce Tik-Sze Li, Chui-Ping Lee, and Wai-Kwong Tang,
\newblock ``Changes in mental health among psychiatric patients during the
  covid-19 pandemic in hong kong—a cross-sectional study,''
\newblock {\em International Journal of Environmental Research and Public
  Health}, vol. 19, no. 3, pp. 1181, 2022.

\bibitem{moe2022risk}
Aubrey~M Moe, Elyse Llamocca, Heather~M Wastler, Danielle~L Steelesmith, Guy
  Brock, Jeffrey~A Bridge, and Cynthia~A Fontanella,
\newblock ``Risk factors for deliberate self-harm and suicide among adolescents
  and young adults with first-episode psychosis,''
\newblock {\em Schizophrenia bulletin}, vol. 48, no. 2, pp. 414--424, 2022.

\bibitem{ahmad2022designing}
Rangina Ahmad, Dominik Siemon, Ulrich Gnewuch, and Susanne Robra-Bissantz,
\newblock ``Designing personality-adaptive conversational agents for mental
  health care,''
\newblock {\em Information Systems Frontiers}, pp. 1--21, 2022.

\bibitem{rezaii2022natural}
Neguine Rezaii, Phillip Wolff, and Bruce~H Price,
\newblock ``Natural language processing in psychiatry: the promises and perils
  of a transformative approach,''
\newblock {\em The British Journal of Psychiatry}, pp. 1--3, 2022.

\bibitem{shum2018eliza}
Heung-Yeung Shum, Xiao-dong He, and Di~Li,
\newblock ``From eliza to xiaoice: challenges and opportunities with social
  chatbots,''
\newblock {\em Frontiers of Information Technology \& Electronic Engineering},
  vol. 19, no. 1, pp. 10--26, 2018.

\bibitem{zemvcik2019brief}
M~Tm{\'a}{\v{s}} ZEM{\v{C}}{\'I}K,
\newblock ``A brief history of chatbots,''
\newblock {\em DEStech Transactions on Computer Science and Engineering}, vol.
  10, 2019.

\bibitem{resnik2015beyond}
Philip Resnik, William Armstrong, Leonardo Claudino, Thang Nguyen, Viet-An
  Nguyen, and Jordan Boyd-Graber,
\newblock ``Beyond lda: exploring supervised topic modeling for
  depression-related language in twitter,''
\newblock in {\em Proceedings of the 2nd workshop on computational linguistics
  and clinical psychology: from linguistic signal to clinical reality}, 2015,
  pp. 99--107.

\bibitem{zeng2012synonym}
Qing~T Zeng, Doug Redd, Thomas Rindflesch, and Jonathan Nebeker,
\newblock ``Synonym, topic model and predicate-based query expansion for
  retrieving clinical documents,''
\newblock in {\em AMIA Annual Symposium Proceedings}. American Medical
  Informatics Association, 2012, vol. 2012, p. 1050.

\bibitem{miao2016neural}
Yishu Miao, Lei Yu, and Phil Blunsom,
\newblock ``Neural variational inference for text processing,''
\newblock in {\em International conference on machine learning}. PMLR, 2016,
  pp. 1727--1736.

\bibitem{miao2017discovering}
Yishu Miao, Edward Grefenstette, and Phil Blunsom,
\newblock ``Discovering discrete latent topics with neural variational
  inference,''
\newblock in {\em International Conference on Machine Learning}. PMLR, 2017,
  pp. 2410--2419.

\bibitem{nan2019topic}
Feng Nan, Ran Ding, Ramesh Nallapati, and Bing Xiang,
\newblock ``Topic modeling with wasserstein autoencoders,''
\newblock in {\em Proceedings of the 57th Annual Meeting of the Association for
  Computational Linguistics}, 2019, pp. 6345--6381.

\bibitem{dieng2020topic}
Adji~B Dieng, Francisco~JR Ruiz, and David~M Blei,
\newblock ``Topic modeling in embedding spaces,''
\newblock {\em Transactions of the Association for Computational Linguistics},
  vol. 8, pp. 439--453, 2020.

\bibitem{wang2020neural}
Rui Wang, Xuemeng Hu, Deyu Zhou, Yulan He, Yuxuan Xiong, Chenchen Ye, and
  Haiyang Xu,
\newblock ``Neural topic modeling with bidirectional adversarial training,''
\newblock in {\em Proceedings of the 58th Annual Meeting of the Association for
  Computational Linguistics}, 2020, pp. 340--350.

\bibitem{mikolov2013distributed}
Tomas Mikolov, Ilya Sutskever, Kai Chen, Greg~S Corrado, and Jeff Dean,
\newblock ``Distributed representations of words and phrases and their
  compositionality,''
\newblock {\em Advances in neural information processing systems}, vol. 26,
  2013.

\bibitem{roder2015exploring}
Michael R{\"o}der, Andreas Both, and Alexander Hinneburg,
\newblock ``Exploring the space of topic coherence measures,''
\newblock in {\em Proceedings of the eighth ACM international conference on Web
  search and data mining}, 2015, pp. 399--408.

\bibitem{mimno2011optimizing}
David Mimno, Hanna Wallach, Edmund Talley, Miriam Leenders, and Andrew
  McCallum,
\newblock ``Optimizing semantic coherence in topic models,''
\newblock in {\em Proceedings of the 2011 conference on empirical methods in
  natural language processing}, 2011, pp. 262--272.

\bibitem{chang2009reading}
Jonathan Chang, Sean Gerrish, Chong Wang, Jordan Boyd-Graber, and David Blei,
\newblock ``Reading tea leaves: How humans interpret topic models,''
\newblock {\em Advances in neural information processing systems}, vol. 22,
  2009.

\bibitem{vrehuuvrek2011gensim}
Radim Rehurek, Petr Sojka, et~al.,
\newblock ``Gensim—statistical semantics in python,''
\newblock {\em Retrieved from genism. org}, 2011.

\bibitem{lin2022deep}
Baihan Lin, Guillermo Cecchi, and Djallel Bouneffouf,
\newblock ``Deep annotation of therapeutic working alliance in psychotherapy,''
\newblock {\em arXiv preprint}, 2022.

\bibitem{lin2020predicting}
Baihan Lin, Djallel Bouneffouf, and Guillermo Cecchi,
\newblock ``Predicting human decision making in psychological tasks with
  recurrent neural networks,''
\newblock {\em arXiv preprint arXiv:2010.11413}, 2020.

\bibitem{lin2022predicting}
Baihan Lin, Djallel Bouneffouf, and Guillermo Cecchi,
\newblock ``Predicting human decision making with lstm,''
\newblock in {\em 2022 International Joint Conference on Neural Networks
  (IJCNN)}. IEEE, 2022.

\bibitem{lin2022reinforcement}
Baihan Lin,
\newblock ``Reinforcement learning and bandits for speech and language
  processing: Tutorial, review and outlook,''
\newblock {\em arXiv preprint arXiv:2210.13623}, 2022.

\bibitem{lin2020story}
Baihan Lin, Guillermo Cecchi, Djallel Bouneffouf, Jenna Reinen, and Irina Rish,
\newblock ``A story of two streams: Reinforcement learning models from human
  behavior and neuropsychiatry,''
\newblock in {\em Proceedings of the 19th International Conference on
  Autonomous Agents and MultiAgent Systems}, 2020, pp. 744--752.

\bibitem{lin2020unified}
Baihan Lin, Guillermo Cecchi, Djallel Bouneffouf, Jenna Reinen, and Irina Rish,
\newblock ``Unified models of human behavioral agents in bandits, contextual
  bandits and rl,''
\newblock {\em arXiv preprint arXiv:2005.04544}, 2020.

\bibitem{lin2019split}
Baihan Lin, Djallel Bouneffouf, and Guillermo Cecchi,
\newblock ``{Split Q Learning: Reinforcement Learning with Two-Stream
  Rewards},''
\newblock in {\em Proceedings of the Twenty-Eighth International Joint
  Conference on Artificial Intelligence, {IJCAI-19}}. AAAI Press, 7 2019, pp.
  6448--6449, International Joint Conferences on Artificial Intelligence
  Organization.

\bibitem{lin2021models}
Baihan Lin, Guillermo Cecchi, Djallel Bouneffouf, Jenna Reinen, and Irina Rish,
\newblock ``Models of human behavioral agents in bandits, contextual bandits
  and rl,''
\newblock in {\em International Workshop on Human Brain and Artificial
  Intelligence}. Springer, 2021, pp. 14--33.

\bibitem{lin2022deep2}
Baihan Lin, Guillermo Cecchi, and Djallel Bouneffouf,
\newblock ``Working alliance transformer for psychotherapy dialogue
  classification,''
\newblock {\em arXiv preprint arXiv:2210.15603}, 2022.

\bibitem{lin2022unsupervised}
Baihan Lin,
\newblock ``Personality effect on psychotherapy outcome: A predictive natural
  language processing framework,''
\newblock {\em arXiv preprint}, 2022.

\bibitem{lin2022knowledge}
Baihan Lin,
\newblock ``Knowledge management system with nlp-assisted annotations: A brief
  survey and outlook,''
\newblock in {\em CIKM Workshops}, 2022.

\end{thebibliography}

\end{document}